\documentclass[letterpaper, 10 pt, conference]{ieeeconf}  % Comment this line out if you need a4paper

\IEEEoverridecommandlockouts                              % This command is only needed if 
\usepackage{graphics} % for pdf, bitmapped graphics files
\usepackage{epsfig} % for postscript graphics files
\usepackage{times} % assumes new font selection scheme installed
\usepackage[T1]{fontenc}
\usepackage[ruled,vlined,linesnumbered]{algorithm2e}

\usepackage{amsmath, graphicx}
\usepackage{balance}
\usepackage{amssymb,amsbsy}
\usepackage{breqn,bbm,xcolor}%\usepackage[dvipsnames]{xcolor}
\usepackage{multirow}
\usepackage[symbol,bottom]{footmisc}
\usepackage{soul}
\usepackage{url}
\usepackage{booktabs} % To thicken table lines
\usepackage{tikz}
\usepackage[export]{adjustbox}
\usepackage{subcaption}
\usepackage{float}
\usepackage{tabularx}
\usepackage{amssymb}

\title{\LARGE \bf
Differentiable Rendering-based Pose Estimation for Surgical Robotic Instruments
}

\author{{Zekai Liang$^{1}$,  Zih-Yun Chiu$^{1}$, Florian Richter$^{1}$, Michael C. Yip$^{1}$}% <-this % stops a space
\thanks{$^{1}$Department of Electrical and Computer Engineering, University of California San Diego, La Jolla, CA 92093 USA. {\tt\small\{z9liang, zchiu, frichter, yip\}@ucsd.edu}}% <-this % stops a space
}

\begin{document}

\maketitle
\thispagestyle{empty}
\pagestyle{empty}

%%%%%%%%%%%%%%%%%%%%%%%%%%%%%%%%%%%%%%%%%%%%%%%%%%%%%%%%%%%%%%%%%%%%%%%%%%%%%%%%
\begin{abstract}

Robot pose estimation is a challenging and crucial task for vision-based surgical robotic automation.
Typical robotic calibration approaches, however, are not applicable to surgical robots, such as the da Vinci Research Kit (dVRK) \cite{kazanzides2014open}, due to joint angle measurement errors from cable-drives and the partially visible kinematic chain.
% These challenges usually lead to the failure of traditional robot pose estimation methods in surgical practice. 
% In recent years, some works have been proposed to tackle surgical tool pose estimation with imprecise joint angle readings. However, many of these methods either overly rely on data-hungry learning-based approaches or require extra sensors like depth, while some are still in need of cumbersome physical setup.
Hence, previous works in surgical robotic automation used tracking algorithms to estimate the pose of the surgical tool in real-time and compensate for the joint angle errors.
However, a big limitation of these previous tracking works is the initialization step which relied on only keypoints and SolvePnP.
In this work, we fully explore the potential of geometric primitives beyond just keypoints with differentiable rendering, cylinders, and construct a versatile pose matching pipeline in a novel pose hypothesis space.
We demonstrate the state-of-the-art performance of our single-shot calibration method with both calibration consistency and real surgical tasks.
As a result, this marker-less calibration approach proves to be a robust and generalizable initialization step for surgical tool tracking.
\end{abstract}

%%%%%%%%%%%%%%%%%%%%%%%%%%%%%%%%%%%%%%%%%%%%%%%%%%%%%%%%%%%%%%%%%%%%%%%%%%%%%%%%
\section{Introduction}

Automation in robot-assisted surgery has been widely explored in the past few years and has been implemented in many real-world surgical tasks, such as automatic suturing \cite{joglekar2023suture, chiu2021bimanual, hari2024stitch}, robot cutting \cite{haiderbhai2024sim2real, tian2023virtual, heiden2021disect}, dissection \cite{liang2024medic} and tensioning \cite{thananjeyan2017multilateral, li2024method}.
Minimal invasive robot surgery with accurate manipulation of the surgical instruments requires precise localization of the robot end-effector especially in vision-based feedback control. Compared to typical robot pose estimation problems, surgical tool pose estimation is inherently more complex and presents unique challenges. One common issue is hysteresis and cable tension brought by cable-driven robot system design \cite{kazanzides2014open, hannaford2012raven}.
Thus, the joints configuration can be very hard to track from robot encoders due to unavoidable cable stretching with tension, which makes the direct joint angle readings errorful and unreliable. 

Under this condition, many state-of-the-art methods have been proposed to address such issues and aim to achieve robust and accurate end-effector localization. Early approaches include \cite{hwang2020applying, seita2018fast} which propose novel physical systems to compensate the imprecision. And later \cite{hwang2020efficiently} designs 3D printed fiducial coordinate frames on the robot end-effector which is tracked by RGBD camera and the calibration result can be obtained from sampled trajectories analysis. However, these approaches rely on tedious physical setup and extra depth sensors, which can be hard to reproduce and apply in real surgical scenarios.

\begin{figure}[t]
    \centerline{\includegraphics[width=1\linewidth,clip=true,trim={0mm 5mm 0mm 0mm}]{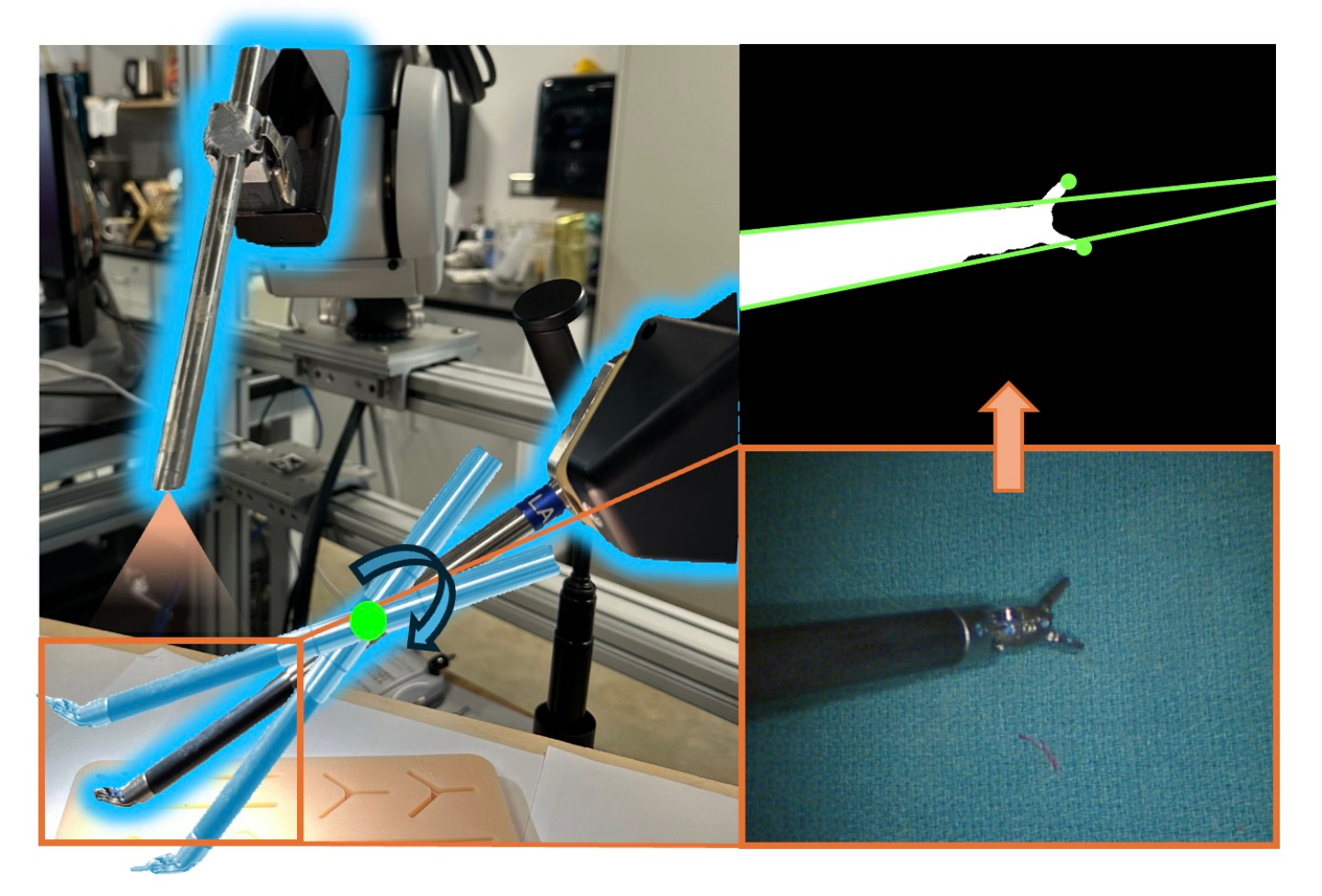}}
    \caption{The motion of the da Vinci surgical robot is constrained by the Remote Center of Motion (RCM). In the endoscopic view, typically only the gripper and insertion shaft are visible. By fully leveraging these visual features, we construct a differentiable rendering-based framework to estimate globally optimal instrument pose. }
    \label{demonstration}
    % \vspace{-0.20in}
    \vspace{-0.14in}
\end{figure}

 % \vspace{-0.20in}
\begin{figure*}[ht]
    \centerline{\includegraphics[width=0.95\textwidth,clip=true,trim={0mm 0mm 0mm 0mm}]{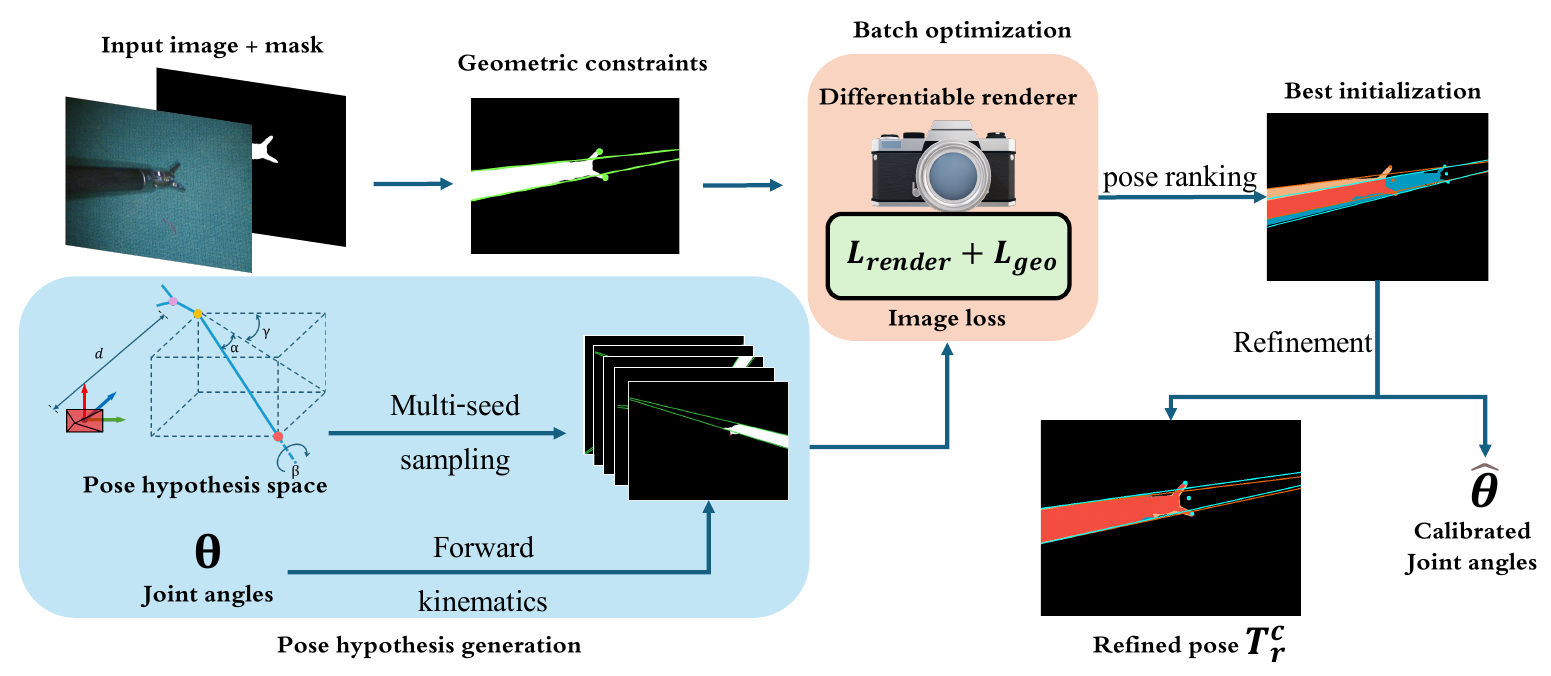}}
    \caption{Our calibration pipeline. We use low resolution image frames, masks and inaccurate joint angles as input. We generate a batch of pose candidates from our pose hypothesis space. Coarse batch optimization is conducted to select the candidate with best converging performance. We further refine the initial pose and output the final estimation with calibrated joint angles.}
    \label{pipeline}
    \vspace{-0.14in}
    
\end{figure*}

More recently, vision-based approaches have drawn more attention due to their convenience and applicability, such as \cite{reiter2014appearance, li2020super, lu2021super} that detect keypoints as visual features and pass them into Particle Filter to track real-time robot pose. \cite{richter2021robotic} first proposes a unified framework that includes geometric features such as cylinder into the visual feature detection. \cite{d2024robust} further upgrades this with a more robust observation model for edge detection. 
 However, in the initialization step, these methods all first detect 2D point features, sometimes even painted on the end-effector, and use a PnP (Perspective-n-Points) solver to obtain the initial camera-to-robot transform. The PnP solver is particularly prone to errors from poor video quality and noise in the keypoint detection.
 And such an initialization step is highly unreliable and sometimes even takes extra physical setup like painting the surgical tool.
% \textcolor{red}{TODO: adding super and KCS, some adjustment of intro and related work}

% \cite{richter2021robotic} first proposes a unified framework that includes geometric features of the dVRK arm into the calibration procedure and integrates it with Particle Filter. Based on this, \cite{d2024robust} further upgrades with a more robust observation model.

% \cite{zhao2015efficient} defines a kinematic remote-center coordinate system (KCS) which absorbs all the error in the estimated transform.

In this work, we present the first rendering-based surgical tool calibration framework, as shown in Fig. \ref{demonstration}, that achieves robust markerless one-shot calibration performance as a superior solution for initial camera-to-robot transform estimation. We integrate geometry primitives with differentiable renderer matching. We adopt a multi-seed start strategy in a robot pose hypothesis space, which significantly improves the efficiency of pose candidate generation. A novel objective loss function with geometric constraints is proposed to tackle the local minima introduced by uneven robot design. We evaluate our method's calibration consistency and apply it in real-world manipulation tasks, showing state-of-the-art single-shot calibration performance. In summary, our contributions are threefold:
\begin{itemize}
    \item The first differentiable rendering-based approach in surgical robot pose estimation.

    \item A pose hypothesis space that accurately parametrizes the da Vinci surgical robot arm orientation, leading to significant improvements in pose sampling strategy.

    \item A novel objective loss function with geometric constraints that tackles local minima and achieves fast matching convergence.
\end{itemize}

% Such as poor video quality, inaccurate joint angle readings, partially visible kinematic chains, and less common robot geometric shapes.

\section{Related Work}

\subsection{Robot pose estimation}
In the past few years, accurately estimating robot pose from visual information has been a crucial task for vision-based control. Traditional methods usually attach physical fiducial markers on the robot end-effector to calibrate the camera parameters \cite{garrido2014aruco,olson2011apriltag}. Such approaches can achieve high accuracy and reliable calibration of camera-to-robot transform but suffer from cumbersome physical setup and lack of flexibility, which requires operators to repeat the whole tedious process once the camera placement is perturbed.

With the rise of Deep Neural Networks (DNN), learning-based approaches have been applied in the marker-less robot pose estimation. Prior work includes \cite{lambrecht2019towards, lee2020camera, lu2022pose, lu2021super} which leverages a keypoint detector to estimate the articulated robot pose with a PnP solver. There also appears another rendering-based category such as \cite{labbe2021single, lu2023image} that demonstrates the potential of the render-match approach in pose estimation. CtRNet \cite{lu2023markerless} combines the keypoints and rendering methods to construct a self-supervised training pipeline that makes training with unlabeled real world data possible. \cite{lu2024ctrnet} further improves to handle more practical and common scenarios. These approaches have made impressive progress in robot pose estimation, whereas their applicability is heavily constrained in surgical robotics due to the complexity that comes together with surgical scenes and unconventional robot system design.

\subsection{Surgical instruments localization}

In surgical scenes, robot links flexibility introduced by cable-driven system design makes it hard to accurately measure joint angles. Early works include measuring cable stretching and friction \cite{miyasaka2015measurement}, learning end-effector off-sets \cite{mahler2014learning, seita2018fast, pastor2013learning}, calibrating the remote center of motion (RCM) \cite{zhong2020hand, zhao2015efficient} and  physically modeling cable-driven system hysteresis \cite{wang2013online}.
Deep learning-based approaches have been explored in works such as  \cite{lu2022pose, reiter2012feature}. \cite{fan2024reinforcement} incorporates reinforcement learning agent with a pre-trained keypoint detector to match each articulated joint.
However, the common Endoscopic Camera Manipulator (ECM) equipped for dVRK robot only provides limited video resolution and suboptimal lighting, which makes the feature detector prone to errors from poor video quality.

\cite{richter2021robotic} first proposes lumped error which estimates cable system measurement error and robot pose simultaneously with particle Filter, which introduces the cylinder as a geometric primitive to define the space position of the insertion shaft. And \cite{d2024robust} further improves the observation model with Neural Networks. However, these methods overly rely on PnP solver to obtain the initial robot pose which is inaccurate and not practical.

% , increasing the robustness in more noisy and blurred scenes. However, both of these methods require a good initialization from PnP solver to have accurate tracking performance. 

\begin{figure}[t]
    \centerline{\includegraphics[width=1\linewidth,clip=true,trim={0mm 0mm 1mm 1mm}]{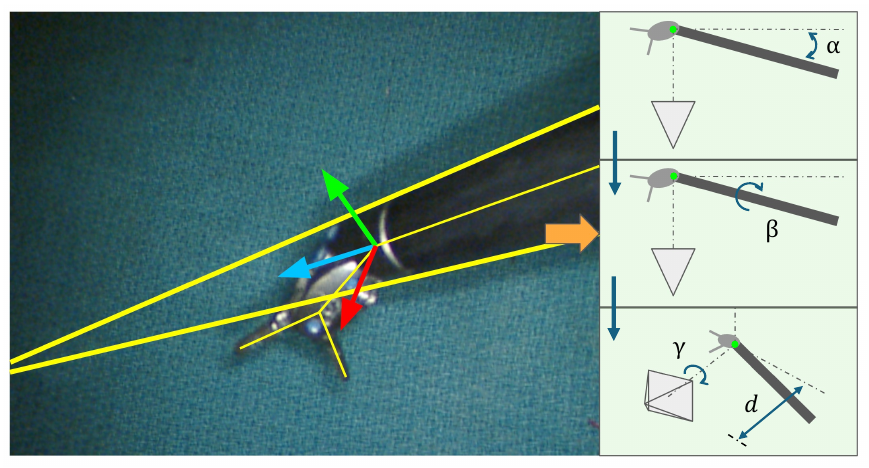}}
    \caption{We parametrize the initial pose candidates into 4 independent Degree of Freedoms, $[\alpha, \beta, \gamma, d]$, covering all potential orientations while leaving out the redundant ones. }
    \label{pose sampling}
    \vspace{-0.14in}
    
\end{figure}

\section{Methodology}

Unlike typical render-match problems explored in prior literature, matching a dVRK robot arm with silhouette images is heavily constrained by its partially visible kinematic chain and the unconventional geometric shape, which consists of two very small tips and a long shaft extending into the camera frame.
These constraints often lead existing approaches to fail due to local minima and poor initialization.
We solve this problem using gradient descent with multi-seed start. Our framework pipeline is shown in Fig. \ref{pipeline} and Algorithm \ref{alg:algorthm}. In the following subsections, we will cover each part and how they contribute to tackling the challenging problem of surgical instrument pose estimation.

\subsection{Multi-seed sampling from pose hypothesis space}

As we discussed in the previous section, partial visibility and the uncommon geometric shape of surgical tools makes the pose initialization an important step for convergence.
% have made many traditional matching-based methods fail in surgical scenes, and such issues are usually caused by poor optimization convergence. Finding a finely initialized start position is a very effective solution to the numerous local minima the optimizer might encounter. 
To efficiently address this problem, we propose a pose hypothesis space based on the robot arm orientation with respect to the endoscopic camera, as shown Fig. \ref{pipeline} and Fig. \ref{pose sampling}.
It simulates the potential robot orientations by positioning the insertion shaft around an anchor point in 3D space. We use this as the sampling space for our multi-seed pose candidate generation.

 We parametrize the initial candidate pose as $[\alpha. \beta, \gamma, d]$, where $\alpha$ and $\gamma$ are rotation angles with respect to the original camera frame, while $\beta$ stands for the rotation around the z axis of the robot local frame. And $d$ is the initial distance from the camera to the robot origin. As such, the initial pose sampling space can be expressed as:

{\small \begin{equation}
    \textbf{T}_{LookAt}(d_i, \alpha_i, \beta_i, \gamma_i)
    =
    \begin{bmatrix}
        \textbf{R}_{LookAt}(\alpha_i, \beta_i, \gamma_i) & \textbf{t}(d_i) \\
        \boldsymbol{0}^\top & 1
    \end{bmatrix}.
\end{equation}}

{\small \begin{equation}
\begin{aligned}
    \textbf{R}_{LookAt}(\alpha_i, \beta_i, \gamma_i) &= 
    \textbf{R}_y(\gamma_i)[\textbf{R}_x(\alpha_i)\textbf{R}_z(\beta)\textbf{R}_x(\alpha_i)^{-1}]\textbf{R}_x(\alpha_i) \\
     &= \textbf{R}_y(\gamma_i)\textbf{R}_x(\alpha_i)\textbf{R}_z(\beta)  
\end{aligned}
\end{equation}}

\begin{equation}
    \textbf{t}(d_i, \alpha_i, \beta_i, \gamma_i) = [0, 0, d_i]
\end{equation}
where $[R_x(\alpha_i)R_z(\beta)R_x(\alpha_i)^{-1}]$ is the rotation around the local z-axis of the robot origin which aligns with the cylinder's normal orientation. In this pose hypothesis space we make an assumption that the insertion shaft is constrained by limited orientation angles relative to the camera frame, as it would not point towards or penetrate the camera plane. We also assume that the cylinder would have another independent Degree of Freedom to rotate along its normal direction in any position in space. 

\begin{figure}[t]
    \centerline{\includegraphics[width=0.9\linewidth,clip=true,trim={0mm 0mm 3mm 7mm}]{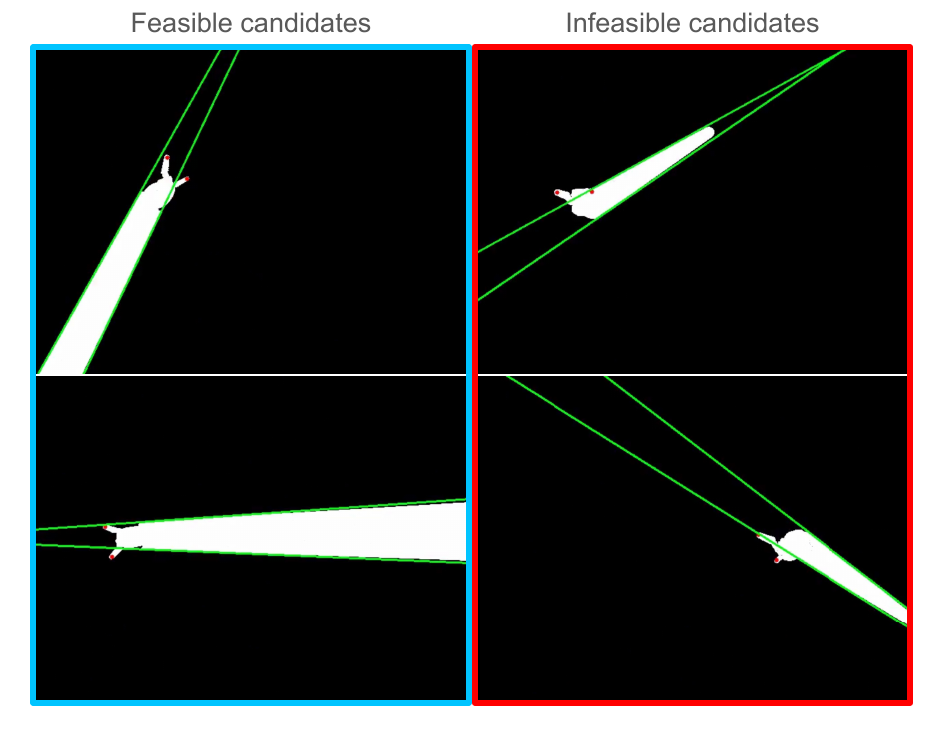}}
    \centering
    \footnotesize
    \hspace{30pt}  Look At Samples   \hspace{30pt}   Random Uniform Samples \hspace{15pt}
    \caption{Our pose hypothesis space focuses on efficiently sampling all the potential feasible poses in real dVRK manipulation scenes while leaving out redundant ones, which greatly boosts the calibration speed and accuracy with limited sample quantity. }
    \label{pose comparison}
    \vspace{-0.14in}
    
\end{figure}

Compared with uniform 6D pose sampling, our sampling space can efficiently avoid infeasible pose candidates and thus makes it possible to find the best starting pose with a small pose candidate quantity, as shown in Fig. \ref{pose comparison}.
We utilize differentiable renderer from Pytorch3D \cite{ravi2020pytorch3d} to generate all the pose candidates by inputting the dVRK robot CAD model and initial poses $\textbf{T}_{LookAt}$, and transforming the mesh vertices $\textbf{v}_{m} \in \mathbb{R}^{n \times 3}$ to camera frame with forward kinematics by
\begin{equation}
    \textbf{v}_{c} = \textbf{T}_{LookAt} \textbf{T}_{m}^{r}(\textbf{q}) \textbf{v}_{m}
\end{equation}
where $\textbf{T}_{m}^{r}$ is the transform from CAD mesh frame to robot frame and $\textbf{v}_{c}$ is the ready-to-render vertices in the camera frame. The binary masks are generated by a silhouette renderer.

\begin{algorithm}[t]
    \small
    \caption{Surgical Tool Parameter Estimation}
    \label{alg:algorthm}
    \KwIn{Image $\mathbb{I}$, initial joint angles $\theta$, link meshes $\mathcal{M}$}
    \KwOut{Camera-to-robot transform $\textbf{T}$, calibrated joint angles $\hat{\theta}$}

    \CommentSty{// Reference mask} \\

    $\mathbb{M}_{\text{ref}} \gets \text{SAM}(\mathbb{I})$ \\

    \CommentSty{// Generate pose candidates} \\

   \For{$i \gets 1$ \KwTo $N$}{
    $(\alpha_i, \gamma_i, \beta_i, d_i) \gets \text{PoseSample}(i)$\;
    $\textbf{T}_i \gets \text{LookAtTransform}(\alpha_i, \gamma_i, \beta_i, d_i)$\;
    }
    $\mathcal{T} \gets \{\textbf{T}_i\}_{i=1}^{N}$

    \CommentSty{// Rank candidates} \\

    \ForEach{$\textbf{T} \in \mathcal{T}$}{
    $\mathbb{S} \gets \text{SilhouetteRenderer}(T, \theta, \mathcal{M})$\;
    $L_{\textbf{T}} \gets \text{ObjectiveLoss}(\mathbb{S}, \mathbb{M}_{\text{ref}})$\;
    }

    \CommentSty{// Get best candidate} \\

    $\textbf{T}_{\text{best}} \gets \underset{\mathbf{T}}{\arg\min}( L)$\;

    \CommentSty{// Refine best candidate} \\

    \ForEach{$i \gets 1$ \KwTo $M$}
    {
        $\mathbb{S}_i \gets \text{SilhouetteRenderer}(\textbf{T}_{\text{best}, i}, \theta, \mathcal{M})$ \\
        $L_i \gets \text{ObjectiveLoss}(\mathbb{S}_i, \mathbb{M}_{\text{ref}})$ \\
        % $\tilde{T}_c^b, \hat{\theta} \gets \text{GradientDescent}(L, T_{\text{best}})$
        $\textbf{T}_{\text{best}} \leftarrow \alpha \frac{\partial L_i}{\partial \textbf{T}_{\text{best}}} + \textbf{T}_{\text{best}}$ \\
        $\hat{\theta} \leftarrow \alpha \frac{\partial L_i}{\partial \theta} + \theta$
    }
    
    \Return $\textbf{T}_{\text{best}}, \hat{\theta}$
       
\end{algorithm}

\subsection{Rendering Loss}

Our objective is to construct a loss function that not only exploits the converging potential of any initial pose in space, but also provide a good standard in pose selection.
We start addressing this problem by applying the Mean Squared Error (MSE) loss for gradient descent:
\begin{equation}
    \mathcal{L}_{\text{MSE}} = \sum_{i=0}^{H-1} \sum_{j=0}^{W-1} \left( \mathbb{S}(i, j) - \mathbb{M}_{\text{ref}}(i, j) \right)^2.
\end{equation}
where $(i, j)$ stands for each pixel in the image, while $\mathbb{M}_{\text{ref}}$ and $\mathbb{S}$ are the reference and rendered silhouette masks, respectively.

Since the MSE does not return an informative gradient if there exists no overlapped region between the two masks, we generate a distance field for the reference mask using  \textit{scikit-fmm} package to propagate the gradient for the entire image:
\begin{equation}
\mathbb{D}_{\text{ref}}(i, j) =
\begin{cases} 
0, & \mathbb{M}_{\text{ref}}(i, j) = 1 \\[8pt]
\frac{\text{dist}(i, j)}{\gamma}, & \mathbb{M}_{\text{ref}}(i, j) = 0
\end{cases}
\end{equation}
\begin{equation}
\mathcal{L}_{\text{dist}} = \sum_{i=0}^{H-1} \sum_{j=0}^{W-1} \mathbb{S}(i, j) \cdot \mathbb{D}_{\text{ref}}(i, j)
\end{equation}
where $D_{\text{ref}}$ is the distance field of the reference mask and $\gamma$ is a decay factor. 
Additionally, we also apply a scale loss which compares the total positive mask pixel to adjust the distance of the robot model :
\begin{equation}
\mathcal{L}_{\text{scale}} = \left\lVert 
\sum_{i=0}^{H-1} \sum_{j=0}^{W-1} \mathbb{S}(i, j) 
- 
\sum_{i=0}^{H-1} \sum_{j=0}^{W-1} \mathbb{M}_{\text{ref}}(i, j)
\right\rVert
\end{equation}
Finally, the rendering loss for image-level silhouette pose matching can be defined as
\begin{equation}
   \mathcal{L}_{\text{render}} =  \lambda_1 \mathcal{L}_{\text{MSE}} + \lambda_2 \mathcal{L}_{\text{dist}} + \lambda_3 \mathcal{L}_{\text{scale}}
\end{equation}
where $\lambda_{1,2,3}$ are scale factors for each term.

\subsection{Geometry Loss}

As shown in the rendering loss, we utilize different conventional binary matching loss functions to achieve the best alignment. However, due to the geometric nature of the surgical tool, the gripper orientation and shaft angle can be greatly perturbed by slight robot pose shifts while the rendering loss might not be able to capture this change. Inspired by our previous work \cite{richter2021robotic} which proposes the cylinder as a crucial geometric primitive in dVRK calibration, we take the two edge lines of the insertion shaft as the geometric constraint to build additional sensitivity for the shaft angle change. 

% while setting the joint angles of end-effectors as optimizable parameters.

\begin{figure*}[ht]
    \centerline{\includegraphics[width=0.98\linewidth,clip=true,trim={0mm 0mm 0mm 0mm}]{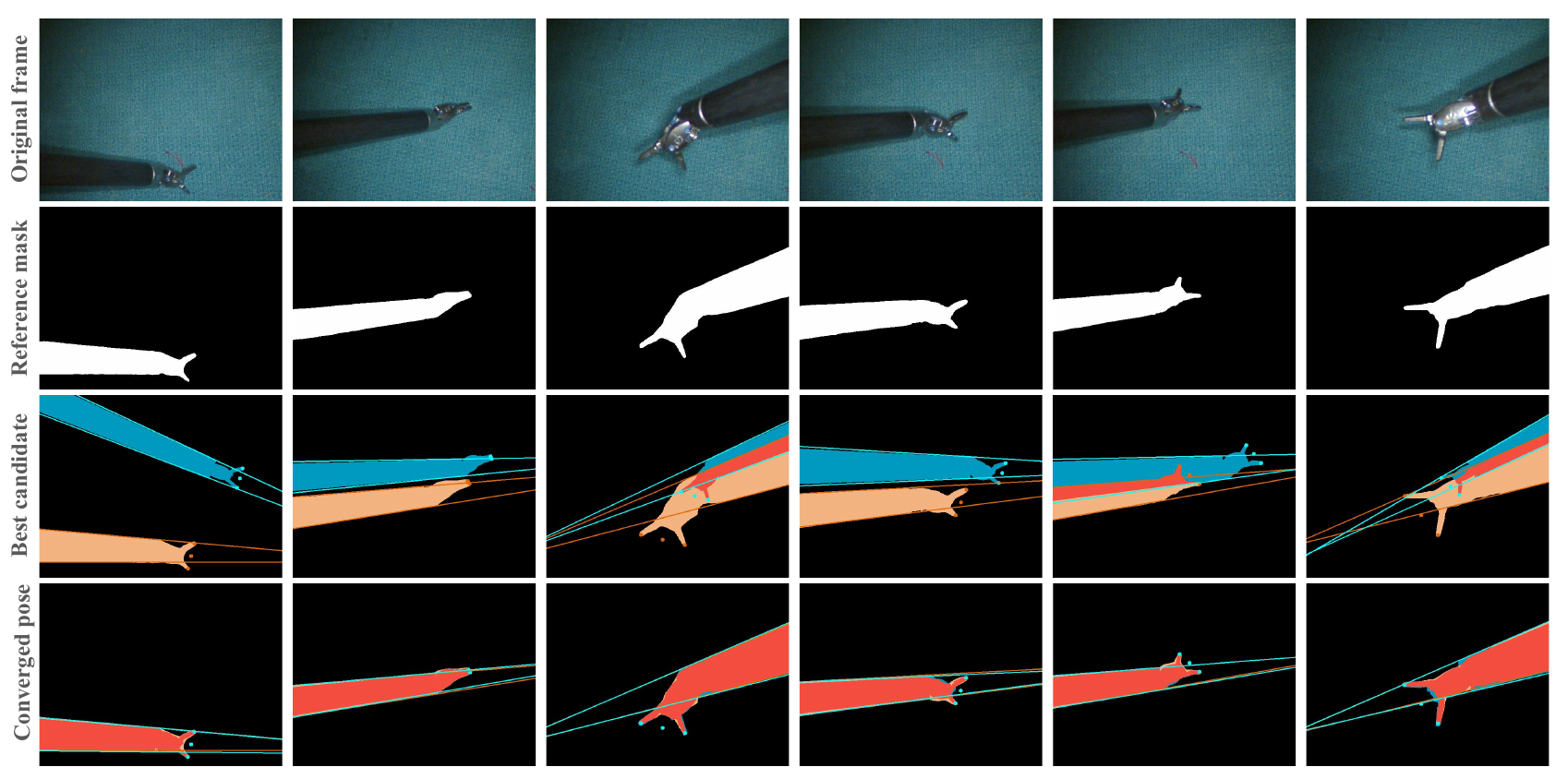}}
    \caption{Quantitative results of our methods applied to real-world images of a dVRK's surgical manipulator. The best candidate of convergence is colored in blue, and the reference mask is colored in orange in the last two columns. We visualize how well the masks and the insertion shaft's edge lines align between the reference and estimated masks. The improved alignment from the third to the fourth columns demonstrates that our method leads to accurate estimation given the visual information.}
    \label{quantitative results}
    \vspace{-0.14in}
\end{figure*}

For any point on the cylinder surface in 3D space, it satisfies:
\begin{equation}
    \begin{cases}
(\mathbf{p} - (\mathbf{p}_0^c+\lambda \mathbf{d}^c)^\top (\mathbf{p} - (\mathbf{p}_0^c+\lambda \mathbf{d}^c)) - r^2 = 0 \\
(\mathbf{d}^c)^\top (\mathbf{p} - (\mathbf{p}_0^c+\lambda \mathbf{d}^c)) = 0
\end{cases}
\end{equation}
where $\mathbf{p}$ is a random surface point, $\mathbf{p}_0^c$ is a point on the cylinder center line and $\mathbf{d}^c$ is the norm of the center line in 3D space. We project the two cylinder edges to the camera frame using a pinhole camera model in the form of:

\begin{equation}
\begin{aligned}
    a_1 X + b_1 Y + c_1 &= 0 \\
    a_2 X + b_2 Y + c_2 &= 0 
\end{aligned}
\label{abc}
\end{equation}

\begin{equation}
    (X, Y) = \left( \frac{u - c_u}{f_x}, \frac{v - c_v}{f_y} \right)
\label{XY}
\end{equation}
where $(u, v)$ is the corresponding pixel coordinate in the image frame, $c_u, c_v, f_X, f_y$ are the camera principal point and focal length, and $a_{1,2}, b_{1,2}, c_{1,2}$ are the normalized line parameters of the cylinder edges projected to the camera plane. These parameters can be calculated using one point $\textbf{p}_0^c = [x_0^c, y_0^c, y_0^c]$ on the center line and the line orientation $\textbf{d}^c = [a^c, b^c, c^c]$ as input:

\begin{equation}
    \small
    \begin{aligned}
    A_{1,2} &=
    \frac{r \left( x_0^c - a^c (\mathbf{p}_0^c)^\top \mathbf{d}^c \right)}
    {\sqrt{(\mathbf{p}_0^c)^\top \mathbf{p}_0^c - (\mathbf{p}_0^c)^\top \mathbf{d}^c - r^2}}
    \pm (c^c y_0^c - b^c z_0^c) 
    \\
    B_{1,2} &=
    \frac{r \left( y_0^c - b^c (\mathbf{p}_0^c)^\top \mathbf{d}^c \right)}
    {\sqrt{(\mathbf{p}_0^c)^\top \mathbf{p}_0^c - (\mathbf{p}_0^c)^\top \mathbf{d}^c - r^2}}
    \pm (a^c z_0^c - c^c x_0^c) 
    \\
    C_{1,2} &=
    \frac{r \left( z_0^c - c^c (\mathbf{p}_0^c)^\top \mathbf{d}^c \right)}
    {\sqrt{(\mathbf{p}_0^c)^\top \mathbf{p}_0^c - (\mathbf{p}_0^c)^\top \mathbf{d}^c - r^2}}
    \pm (b^c x_0^c - a^c y_0^c)
    \end{aligned}
    \label{ABC}
\end{equation}
where $r$ is the radius of the cylinder. Combining \eqref{abc} \eqref{XY} \eqref{ABC}  we can obtain the normalized parametric edge lines on the image plane:
\begin{equation}
    \bar{a}_{1,2} u + \bar{b}_{1,2}v = 1
\end{equation}
The detailed derivation can be referenced in \cite{richter2021robotic}.

To finalize the loss, we further convert the standard form of the line into polar form
\begin{equation}
x \cos\theta + y \sin\theta = \rho
\end{equation}
where
\begin{equation}
\begin{aligned}
    \rho &=  \frac{1}{\sqrt{a^2 + b^2}} 
    \\
    \theta &= \tan^{-1} \left( \frac{b}{a} \right)
\end{aligned}
\end{equation}
We represent the difference between two parametric lines as:
\begin{equation}
    d = \min \left( |\theta_1 - \theta_2|, \pi - |\theta_1 - \theta_2| \right) + \gamma|  \rho_1 - \rho_2 |
\end{equation}
where $\gamma$ is the coefficient for distance $\rho$. As the reference and projected cylinder edges are not pre-paired, we construct our loss function by iterating both cases plus the mean value:

 \begin{equation}
    % \small
    \begin{aligned}
     \mathcal{L}_{\text{cylinder}} = \min((d_{r_1, p_1}+d_{r_2, p_2}),
     \\ 
     (d_{r_1, p_2}+d_{r_2, p_1}))
     + d_{\bar{r}, \bar{p}}
     \end{aligned}
 \end{equation}
 where $r$ is the reference and $p$ is the projection. We detect the cylinder edges using the Canny edge detector \cite{canny1986computational} on binary masks and extracting the two longest edges with the Hough Transform \cite{matas2000robust}.

Besides adopting the cylinder loss to upscale the shaft orientation difference, we also apply a keypoints loss that constrains the gripper pose by placing two keypoints at the end of each tip:

\begin{equation}
\begin{aligned}
    \mathcal{L}_{\text{kpt}} = \min(\sum\|\textbf{p}_{1,2}^{r} - \textbf{p}_{1,2}^{p}\|, \sum\|\textbf{p}_{2,1}^{r} - \textbf{p}_{1,2}^{p}\|) 
    \\
    + \|\textbf{p}_{mean}^{r} - \textbf{p}_{mean}^{p}\|
\end{aligned}
\end{equation}
where $\textbf{p}^{p}$ and $\textbf{p}^{r}$ are projected and reference keypoints, respectively. we calculate the point pairs' position difference with the same association strategy as cylinder edges. We express our geometric loss function with cylinder and keypoint constraints as
\begin{equation}
   \mathcal{L}_{\text{geo}} = \lambda_4 \mathcal{L}_{\text{cylinder}} + \lambda_5 \mathcal{L}_{\text{kpt}}
\end{equation}
With all the constraints we have explored above, our final loss function can be constructed as
\begin{equation}
    \mathcal{L} =  \mathcal{L}_{\text{render}} + \mathcal{L}_{\text{geo}}
\end{equation}
This objective loss function not only helps avoid local minima that can lead to suboptimal convergence, but also provides a robust criterion for pose ranking. During pose candidate selection, it enables rapid convergence under coarse batch optimization with few iterations, sorting out the candidate best aligned from both rendering and geometric aspects. In the subsequent pose refinement stage, the loss function supports further improvement by converging over additional iteration steps.

\section{Experiments and Results}

\subsection{Implementation details}
The calibration framework is implemented in Pytorch with a NVIDIA RTX 3090 GPU. The reference masks are generated using Segment Anything 2 \cite{ravi2024sam}. We use the Adam optimizer \cite{kingma2014adam} for all the gradient propagation. In pose ranking step, we sampled 500 pose candidates for each frame with image resolution $640 \times 480$. We set 100 iteration steps for pose candidate selection with $0.003$ learning rate, and 1000 iteration steps for refinement with $0.0005$ learning rate. In the refinement step,  we implement Gaussian noise perturbation and coordinate descent for better convergence. During the optimization, the visible last three joint angles of the dVRK robot are set as optimizable parameters: Wrist Pitch, Wrist Yaw, and Jaw Angle.

We first show the qualitative calibration performance in Fig. \ref{quantitative results}, including the best pose candidate and final convergence, by visualizing the alignment of geometric features. We test our method on frames with various end-effector poses and mask quality. As shown in the results, the multi-seed start approach successfully transforms the random 6D pose matching into a much easier translation-focused alignment problem by selecting the pose candidate with the most converging potential. 
To evaluate the robustness and accuracy of our approach, we further conduct experiments on calibration consistency and real-world manipulation tasks.

% In the refinement we conduct coordinate decent on rotation and translation in turns with 3 and 1 steps respectively.

% For each single image, the pipeline takes approximately 10 minues to execute.

\begin{figure}[t]
    \centerline{\includegraphics[width=1\linewidth,clip=true,trim={0mm 0mm 0mm 0mm}]{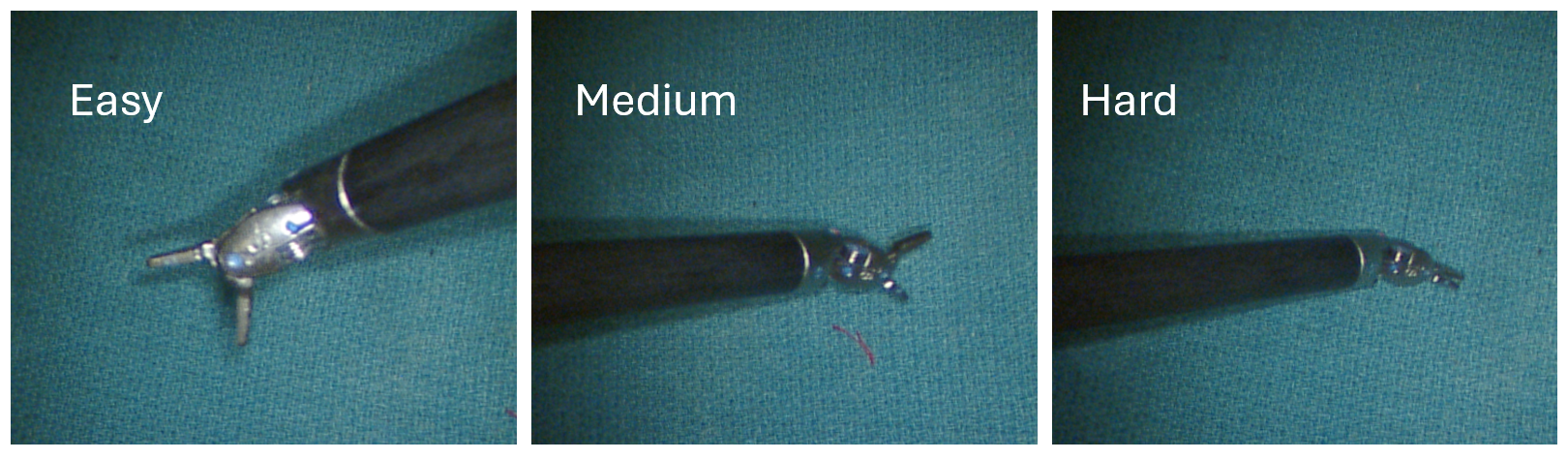}}
    \caption{We collect datasets from videos of a real dVRK robot and set three subsets based on their calibration difficulty. Each subset has 10 frames. }
    \label{consistency difficulty}
    \vspace{-0.14in}
\end{figure}

\subsection{Real world calibration consistency}

% \textcolor{red}{How to localize the RCM, and math?}
% \textcolor{red}{How is the performance of RCM localization in terms of Consistency}

% \textcolor{red}{One figure compare the RCM localization std, the comparison could be PnP from the previous paper.}

\begin{figure}[t]
    \centerline{\includegraphics[width=1\linewidth,clip=true,trim={0mm 0mm 0mm 0mm}]{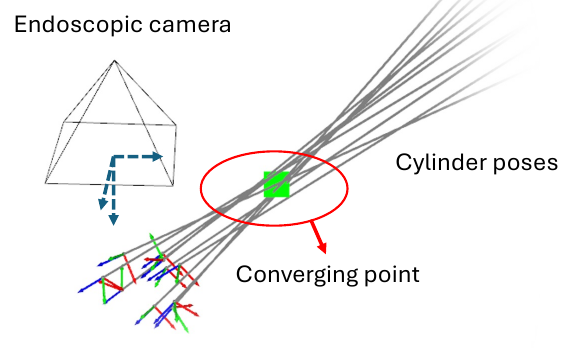}}
    \caption{dVRK robot's motion orbits around the RCM located on the insertion shaft. We locate the converging point from a bunch of estimated cylinder poses.}
    \label{RCM}
    % \vspace{-0.14in}
\end{figure}

We evaluate the calibration consistency by measuring the converging performance of calibrated insertion shafts as a PSM rotates around a fixed virtual point in space, RCM. The experiment is illustrated in Fig. \ref{RCM}.

\begin{table}[t]
    \centering
    \begin{tabular}{l p{1.8cm} p{1.8cm} p{1.8cm} }
    \toprule
    
        Method & Easy & Medium & Hard \\
        \midrule
        
       PnP & 0.05456 & 0.06173 & 0.06761 \\
        Ours & \textbf{0.00046} & \textbf{0.00243} & \textbf{0.00253} \\
       
    \bottomrule    
    \end{tabular}
    \caption{We calculate the standard deviation of converging point distance, in comparison with Perspective-n-Point(PnP)-based approach. The results are in meters (m).}
    \label{consistency}
    \vspace{-0.17in}
\end{table}

\begin{figure*}[ht]
    \centerline{\includegraphics[width=1\linewidth,clip=true,trim={0mm 6mm 0mm 6mm}]{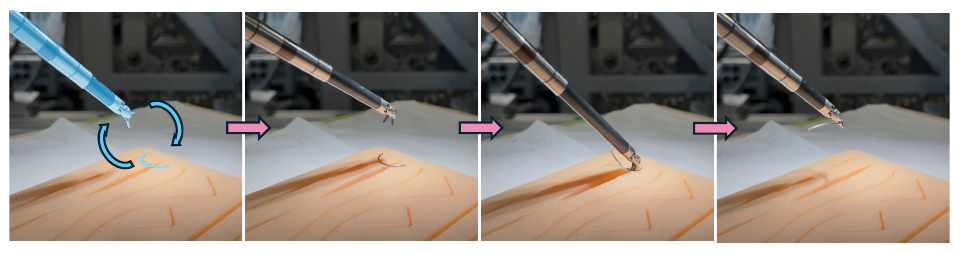}}
    \caption{Experiment setup for medical tools picking with calibrated robot pose. We conduct the picking trials with various initial object and end-effector poses. The picking success rate is recorded with respect to three distance levels}
    \label{setup}
    \vspace{-0.14in}
\end{figure*}

For each estimated camera-to-robot transform, we can obtain the origin location and direction of the insertion shaft. Thus we can locate the converging point by obtaining the closest point in space that has the minimum sum distance to the cylinders. This closest point relative to all the cylinders can be expressed as
\begin{equation}
    \mathbf{x} = \arg \min_{\mathbf{x} \in \mathbb{R}^3} \sum_{i=1}^{N} \left\| (\mathbf{x} - \mathbf{p}_i) - (\mathbf{d}_i^T (\mathbf{x} - \mathbf{p}_i)) \mathbf{d}_i \right\|^2
\end{equation}
where $i$ is the index of the cylinder, $\mathbf{x} \in \mathbb{R}^3$ is the converging point in 3D space, and $d_i$ and $p_i$ are the direction and origin of the cylinder, respectively. Thus, the converging point is the solution to the equation:
\begin{equation}
\mathbf{x}^* = \left( N \mathbf{I} - \sum_{i=1}^{N} \mathbf{d}_i \mathbf{d}_i^T \right)^{-1} \sum_{i=1}^{N} \left( \mathbf{I} - \mathbf{d}_i \mathbf{d}_i^T \right) \mathbf{p}_i
\end{equation}
where $N$ is the number of cylinders we use for calculation. The standard deviation of the distances from this point to each cylinder can be obtained as  
\begin{equation}
    \sigma = \sqrt{\frac{1}{N} \sum_{i=1}^{N} (\|\mathbf{p_i}-\mathbf{x}^*\| - \|\mathbf{\bar{p}}-\mathbf{x}^*\|)^2}
\end{equation}

We evaluate our method on the dataset collected from a real dVRK robot, which includes three difficulty levels (easy, medium, and hard). We measure the standard deviation of the distance distribution and compare it against the PnP solver used in previous studies \cite{richter2021robotic, d2024robust}, as shown in TABLE \ref{consistency}. While the PnP solver is affected by low video resolution and manual point-association issues, our method achieves significantly better calibration consistency without relying on colored physical point labeling, which makes our approach highly generalizable and robust.

\begin{figure}[ht]
    \centerline{\includegraphics[width=1\linewidth,clip=true,trim={0mm 0mm 0mm 0mm}]{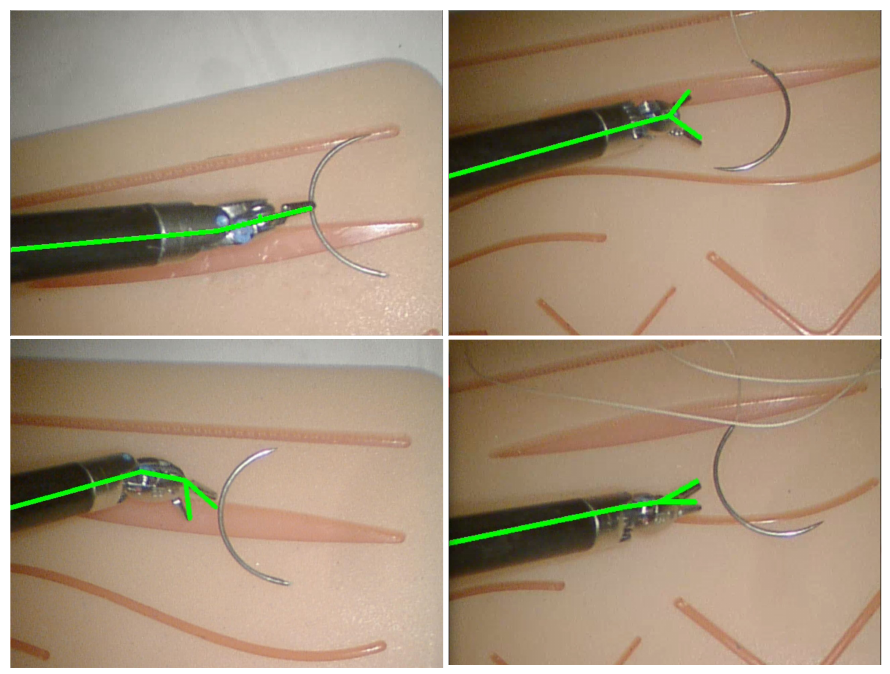}}
    \caption{Projected end-effector pose as green skeleton during suture needle pick up experiment.}
    \label{skeleton}
    \vspace{-0.14in}
\end{figure}

\subsection{Open-loop manipulation with single-shot calibration}

% \textcolor{red}{how to design the real world experiments?}
% \textcolor{red}{moving to target, measure the distance error.}
% \textcolor{red}{pinching objects, measure the success rate. 5 time calibration, 10 time pinching attempt}
% \textcolor{red}{keep it simple}

To better demonstrate the accuracy and applicability of our pose estimation method, we apply it to real-world surgical manipulation tasks with only the initial calibration result. Here, we evaluate the robot's success rate of picking up two surgical tools - a gauze pad and a suture needle - from different levels of distance to the objects. To examine the calibration quality specifically, we carry out open-loop grasps for randomized initial starting poses of the PSM and the object. Our experiment setup is shown in Fig. \ref{setup}. Down-scaled endoscopic images with a resolution of $640 \times 480$ are used for both robot and object pose estimation. A PSM of the dVRK is equipped with a large needle driver that follows the generated motions: \textit{moving to above the object} $\rightarrow$ \textit{approaching and picking up} $\rightarrow$ \textit{dropping the object and resetting the pose}. 

We run the experiments with a gauze pad of size 76 mm $\times$ 76 mm and a needle of radius 1.146 cm. To obtain the gauze-picking pose within the camera frame, we place an ArUco marker on the gauze pad and compute the pose using corner detection and a PnP solver from OpenCV. When experimenting with the suture needle picking, we run prior real-time needle tracking methods~\cite{chiu2022markerless} to obtain the needle-picking pose.

We evaluate the manipulation results by recording the success rate of multiple picking attempts, as shown in TABLE \ref{picking-comparison}. When picking up both objects, the robot starts from three different distance ranges. The near distance is around 1 cm, while the medium and far distances are about 3-4 cm and 7 cm, respectively.

\begin{table}[t]
    \centering
    \begin{tabular}{l || c c c c }
    \toprule
      & \multicolumn{2}{c}{Gauze } & \multicolumn{2}{c}{Needle}\\
    \cmidrule(lr){2-5} 
    &  Success / Trials & Rate &  Success / Trials & Rate \\
    \midrule
    Near    & 11 / 11 & 100.0\% & 8 / 10 & 80.0\% \\
    Medium  & 9 / 10 & 90.0\% & 7 / 11 & 63.4\% \\
    Far    & 10 / 10 & 100.0\% & 5 / 12 & 41.6\%  \\
    \bottomrule
    \end{tabular}
    \caption{Picking success rate comparison for two different objects at three distance ranges.}
    \label{picking-comparison}
    \vspace{-0.14in}
\end{table}

As shown in TABLE \ref{picking-comparison}, the success rate of picking gauze pads—items with relatively large surface areas—approaches 100 \% across all three distance ranges. In suture needle picking, which has a greater demand for precision, the success rate remains high at shorter distances and gradually decreases as the needle gets further. Note that as we run open-loop control for object picking, the imprecision of dVRK's cable-driven joint encoder readings can introduce accumulative pose error when approaching the target, where cables get stretched more. Nevertheless, by projecting the calibrated robot pose in Fig. \ref{skeleton} as a green skeleton, we show excellent alignment between the initial calibrated pose and the original image. This demonstrates that our calibration method greatly contributes to the picking accuracy when cable stretching does not significantly affect the control precision.

% \textcolor{red}{still figuring out how to polish this part}

% \textcolor{red}{3 distances from the calibrated location of the PSM}

% \subsubsection{Bimanual Suture Needle Hand-off}
% \textcolor{red}{3 distances from the calibrated locations of BOTH PSMs}

% \subsubsection{If hand-off or needle picking would be too hard, optionally switch to easier object picking / distance error measurement. Should finish within 2 days.} 

% \begin{figure}[ht]
%     \centerline{\includegraphics[width=1\linewidth,clip=true,trim={0mm 0mm 0mm 0mm}]{figures/skeleton.pdf}}
%     \caption{Projected green robot  skeleton on the real-time video. The calibration result proves to be aligned very well within close range of calibrated position. }
%     \label{setup}
%     % \vspace{-0.14in}
% \end{figure}

\section{Conclusion}
In this work, we propose the first rendering-based single-shot pose estimation framework for surgical robots that achieves accurate and markerless calibration performance. We demonstrate that our method is capable of solving the common local minima and inaccurate joint reading issues by incorporating differentiable rendering with multi-seed start matching. Our loss function is proved to be effective in both pose sampling and matching convergence. By evaluating our method on real manipulation tasks, we show that our method is significantly better than previous single-shot calibration approach used to initialize real-time trackers \cite{li2020super, lu2021super,  richter2021robotic}.
In future work, we intend to incorporate our single-shot approach into active tracking solutions which compensate for joint angle errors in real-time for surgical task automation.
% Ultimately, this work helps to introduce the differentiable rendering approach into surgical robots for the first time,  and could be further improved to handle real-time calibration errors in our future work.


%%%%%%%%%%%%%%%%%%%%%%%%%%%%%%%%%%%%%%%%%%%%%%%%%%%%%%%%%%%%%%%%%%%%%%%%%%%%%%%%

%%%%%%%%%%%%%%%%%%%%%%%%%%%%%%%%%%%%%%%%%%%%%%%%%%%%%%%%%%%%%%%%%%%%%%%%%%%%%%%%

%%%%%%%%%%%%%%%%%%%%%%%%%%%%%%%%%%%%%%%%%%%%%%%%%%%%%%%%%%%%%%%%%%%%%%%%%%%%%%%%

\section*{ACKNOWLEDGMENT}
Special thanks to our lab mates Elizabeth Peiros, Neelay Joglekar and Xiao Liang from UCSD Advanced Robotics and Controls Lab for their insightful discussions and hardware setup assistance.

\balance
\bibliographystyle{ieeetr}
\bibliography{references}

\end{document}